# FRAUNHOFER INSTITUTE FOR EXPERIMENTAL SOFTWARE ENGINEERING IESE

# TOWARDS IDENTIFYING AND MANAGING SOURCES OF UNCERTAINTY IN AI AND MACHINE LEARNING MODELS – AN OVERVIEW

MICHAEL KLÄS

# UNCERTAINTY IN AI AND MACHINE LEARNING – AN OVERVIEW

## Abstract

Quantifying and managing uncertainties that occur when data-driven models such as those provided by AI and machine learning methods are applied is crucial. This whitepaper provides a brief motivation and first overview of the state of the art in identifying and quantifying sources of uncertainty for data-driven components as well as means for analyzing their impact.

## Motivation

Data-driven models (Solomatine & Ostfeld, 2008), (Solomatine, See, & Abrahart, 2009), such as those provided by the application of AI and machine learning, are becoming components of increasing importance for complex software-intensive systems. In particular, embedded systems that collaborate in an open context need to process various kinds of sensor input to recognize and interpret their situation in order to handle changes in their environment and collaborate with previously unknown agents. Unlike traditionally engineered software components, which are developed by software engineers who define their functional behavior using code or models, the behavior of data-driven components is automatically generalized by algorithms from a given data sample.

As a consequence, the functional behavior expected from data-driven components can only be specified in part on their intended domain, and we cannot assure that they will behave as expected in all cases. Moreover, their processing structure is usually difficult to trace and validate by humans because this structure rarely follows human intuition but is generated to provide the algorithmically generalized input-output relationship in an effective manner. Prominent representatives of models used by data-driven components are artificial neural networks and support vector machines (Russell & Norvig, 2016).

Since data-driven models are an important source of uncertainty in embedded systems that collaborate in an open context, the uncertainty they introduce has to be appropriately understood and managed during design time and runtime.

Previous work (Kläs & Vollmer, 2018) proposes separating the sources of uncertainty in data-driven components into three major classes, distinguishing between uncertainty caused by limitations in terms of model fit, data quality, and scope compliance. Whereas model fit focuses on the inherent uncertainty in data-driven models, data quality covers the additional uncertainty caused by their application to input data obtained in suboptimal conditions and scope compliance covers situations where the model is likely applied outside the scope for which it was trained and validated.

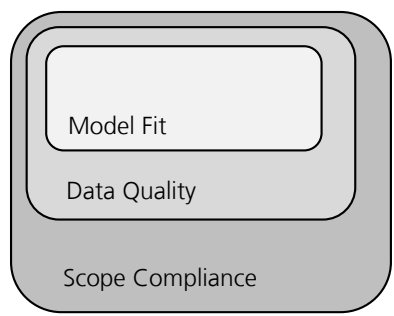

Uncertainty caused by (inherent)
limitations of the learned model

*Additional*
Uncertainty caused by data quality
limitations during model application

*Additional*
Uncertainty caused by mismatch between
target/test context and application context

*Fig. 1: Onion layer model of uncertainty in data-driven model application outcomes (Kläs & Vollmer, 2018)*

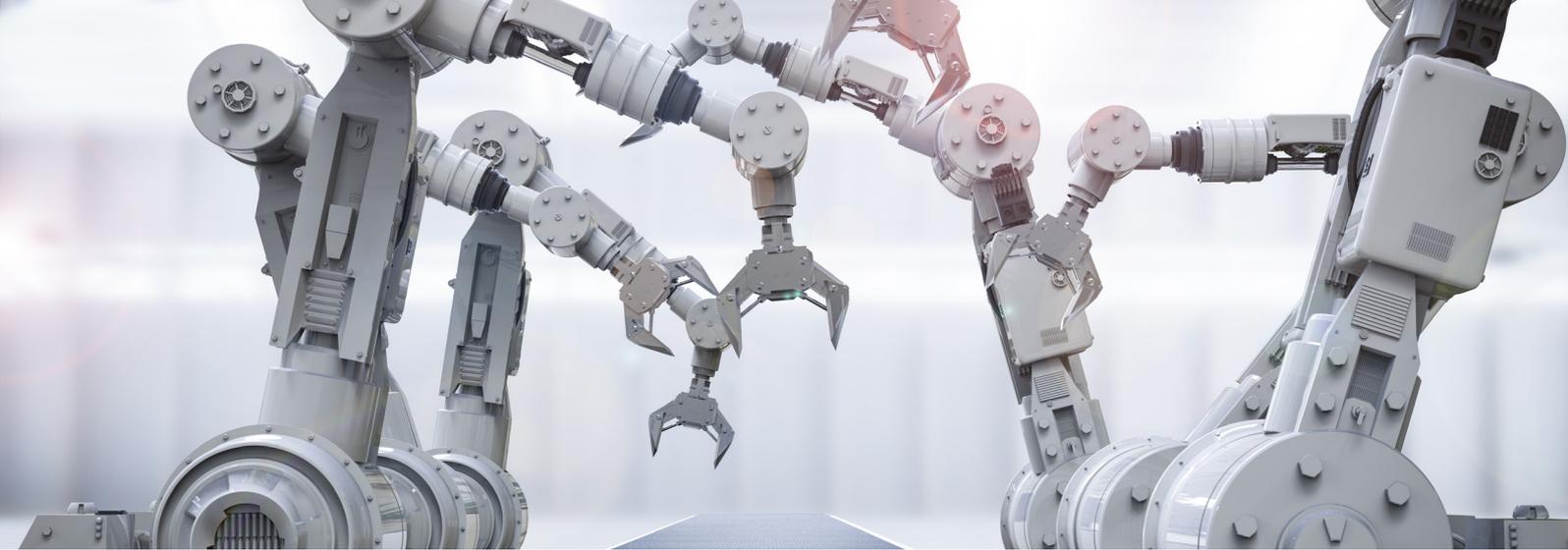

The next sections will provide a short overview of the state of the art in identifying and quantifying sources of uncertainty in data-driven components as well as means for analyzing their impact.

## Model Fit

Model fit uncertainty of a data-driven model is characterized by the degree of error in its outcomes (Kläs & Vollmer, 2018). Thus, model fit uncertainty is related to a concept of model performance commonly named accuracy. Depending on the intended usage of the model, the cost of the model errors, the scale type (e.g., binary, nominal, ordinal, numeric), and the outcome distribution (e.g., degree of balancing), a variety of error measures exist that can be applied to quantify model performance (Japkowicz & Shah, 2011) (Sokolova, Japkowicz, & Szpakowicz, 2006) (Fröhling, 2018). Since the true error of a data-driven model cannot be determined on a data sample, a number of error estimation methods are applied in practice, with the most prominent one being cross-validation (Witten, Frank, Hall, & C., 2016). In order to demonstrate a certain level of confidence in the error estimate, statistical techniques can be applied (e.g., statistical tests and confidence intervals) (Hedderich & Sachs, 2018).

Independent of the specific outcome scale type, the error in a data-driven model regarding unseen data can be decomposed into bias, variance, and irreducible error (Domingos, 2000). Whereas irreducible errors are independent of a specific model, model bias and variance are influenced by the available training data and the selected modeling approach. For example, increasing the amount of training data can reduce model variance, while increasing the complexity of the model can reduce bias. However, because increasing model complexity also increases variance, the sweet spot is targeted by balancing both aspects in order to avoid underfitting (high bias) as well as overfitting (high variance) during model development.

## Data Quality

In realistic settings, data is limited in its accuracy and potentially affected by various kinds of quality issues; therefore, data-driven models are not applied under optimal conditions. For example, the accuracy of a recognition model in identifying a specific object depends on a number of quality characteristics of the input image, such as resolution, light conditions, focus, etc. The delta between the level of uncertainty that can be explained by model fit and the actual level of uncertainty observed in a test situation can be attributed to data quality (Kläs & Vollmer, 2018). Distinguishing between uncertainty related to model fit and uncertainty related to data quality can therefore help to explain variation in the accuracy of the model outcomes within a test dataset and thus provide more specific uncertainty estimates for a concrete application.

Uncertainty estimates are usually expressed by probabilities in the case of categorical outcomes and by prediction intervals together with a confidence level or probability distributions in the case of numerical outcomes (Armstrong, 2001). As the performance of data-driven models can be estimated based on a test dataset, the quality of related uncertainty estimates can also be determined. However, unlike specific model outcomes, which can be evaluated with regard to their individual accuracy, specific uncertainty estimates cannot be evaluated individually. Instead, related evaluation statistics are used to investigate how well the actually measured errors in a set of outcomes are aligned with their corresponding uncertainty estimates. In the case of numerical outcomes, a frequently used evaluation statistic for this purpose is the hit rate (Jorgensen, 2005).

Forward uncertainty propagation is a common approach for dealing with uncertain model input in the context of computation and simulation and is covered by various methods (Lee & Chen, 2009). Founded on the propagation of error theory, forward uncertainty propagation methods consider probability distributions instead of concrete values as model input to

# UNCERTAINTY IN AI AND MACHINE LEARNING – AN OVERVIEW

express corresponding uncertainty. Using techniques such as Monte Carlo simulation, they analyze the resulting uncertainty in the outcome of the model by computing an outcome probability distribution based on specific input probability distributions. In the context of data-driven models, forward uncertainty propagation methods appear to be applicable in cases where the input is provided as structured data and a corresponding probability distribution can be approximated; for example, if the input data consists of values provided by calibrated sensors with known measurement errors. However, computation time requirements may limit their application in design time analysis.

Some classes of data-driven models such as decision trees (Safavian & Landgrebe, 1991) are usually created by modeling techniques that, by default, provide an uncertainty estimate together with their categorical outcome. These estimates are commonly provided as probability values that are derived based on the data used to build the model (cf. Fig. 2).

For further modeling techniques, revisions have been proposed to provide uncertainty estimates (e.g., for some neural network (Khosravi, Nahavandi, Creighton, & Atiya, 2011), deep learning (Gal, 2016) (McAllister, et al., 2017), hybrid (Kläs, et al., 2008), and analogy techniques (Angelis & Stamelos, 2000)). Moreover, some meta-techniques are available that can be applied on top of a variety of existing modeling techniques to obtain uncertainty estimates. Some of them use machine learning techniques for this purpose (e.g., a combination of fuzzy c-means clustering or decision trees with linear regression) (Shrestha & Solomatine, 2006) (Solomatine & Shrestha, 2009). If the applied modeling technique includes means to provide uncertainty estimates, cases need to be carefully differentiated into those where the uncertainty estimates are based on data used during model development and those where estimates are derived based on a previously unseen test dataset. If estimates are based on data used during model development, they may suffer from overfitting, which may lead to overconfident uncertainty estimates when applied to new unseen data. In this case, evaluation strategies need to be defined that do not only evaluate the quality of the data-driven model with respect to its outcomes but also the corresponding uncertainty estimates.

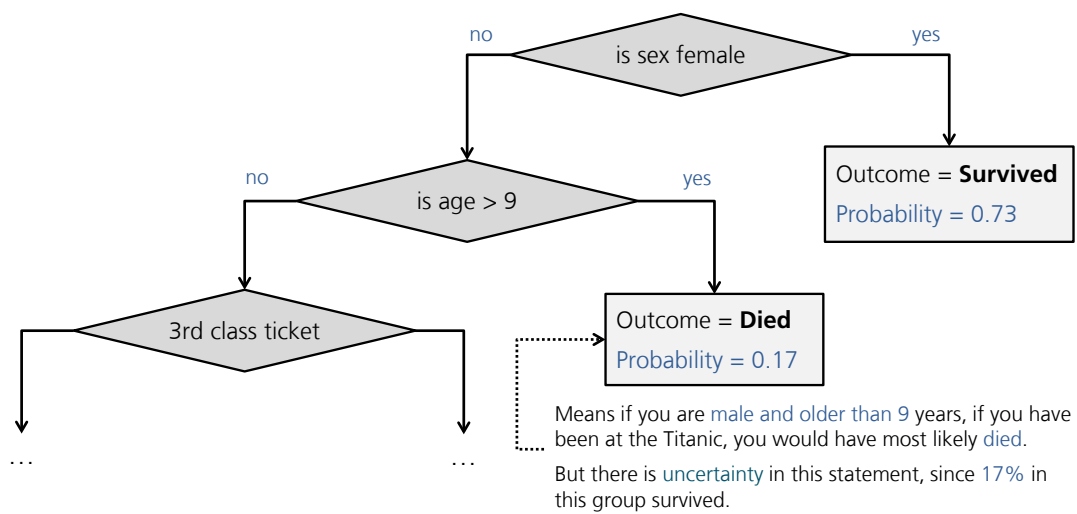

Fig. 2: Decision tree providing not only a binary outcome, but also the degree of uncertainty

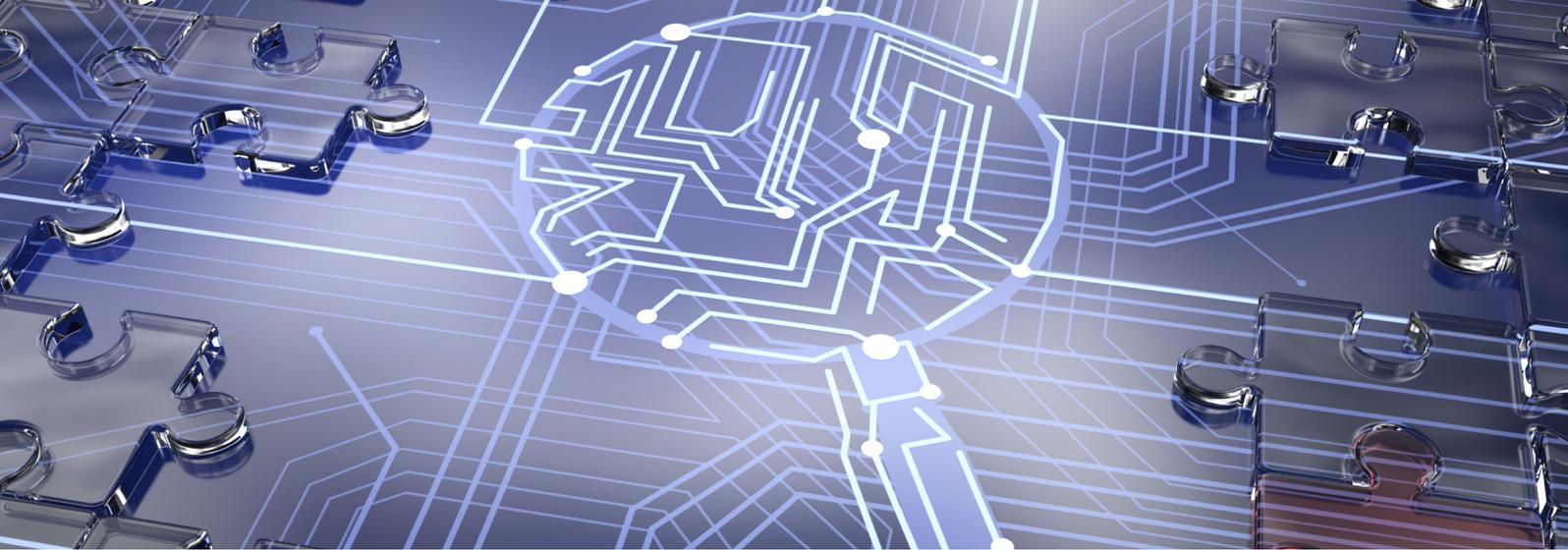

A major drawback of integrating uncertainty estimates directly into data-driven models is that providing realistic uncertainty estimates and providing accurate outcomes does not necessarily require the same inputs. The reason is that information that allows determining situations with high inaccuracy in the current input can help to provide more accurate uncertainty estimates; however, these usually do not support the data-driven model in providing outcomes that are more accurate. For example, the information that the resolution of a specific image is low does not contribute to the recognition of a given object in this image, but may indicate that there is a higher degree of uncertainty in the outcome of the applied object recognition model. Many modeling techniques will therefore eliminate this kind of information during model building in order to reduce variance; however, this may reduce the accuracy of the provided uncertainty estimates. This limitation may be addressed by modeling outcome and uncertainty in two separate models (Solomatine & Shrestha, 2009) or by training one model following a multi-objective approach (Sarro, Petrozziello, & Harman, 2016).

Since the relevant input for uncertainty estimates does not necessarily correspond to the input that is relevant for a data-driven model, data quality models can help to identify and quantify data quality factors that allow increasing the accuracy of uncertainty estimates. Most existing data quality models and standards, such as ISO/IEC 25012 (ISO/IEC, 2008), are, however, defined from the perspective of a human decision maker regarding the characterization and evaluation of data quality. Moreover, existing quality models largely focus on the quality of structured data, missing measures that are applicable to unstructured data such as images, video recordings, or natural language, which are all common inputs in current data-driven models (Kläs, Putz, & Lutz, 2016).

A relevant challenge in determining the impact of data quality on the uncertainty in the outcome of a data-driven model is that data points with specific data quality characteristics (i.e., the ones that can have a large impact on the accuracy of the model outcome) may be sparsely available in a representative data sample and difficult to collect in practice. Data augmentation techniques (Wong, Gatt, Stamatescu, & McDonnell, 2016) (Baird, 1992) may provide means for generating synthetic (i.e., artificial but realistic) data. For example, Generative Adversarial Networks (Goodfellow, et al., 2014) have been shown to make it possible to augment existing images with specific weather and lightning conditions (Liu, Breuel, & Kautz, 2017) (Luan, Paris, Shechtman, & Bala, 2017).

# UNCERTAINTY IN AI AND MACHINE LEARNING – AN OVERVIEW

## Scope Compliance

Even if both model-fit- and data-quality-related uncertainty is managed, the outcome of a data-driven model can become unreliable due to the fact that the model is applied in a setting for which it was not intended.

Data-driven models are created with a specific application scope in mind. An application scope can be defined as a (potentially infinite) set of entities or events satisfying a set of common properties and can thus be considered as a statistical population. In statistics, populations are usually characterized by temporal, spatial, and factual aspects. Using traffic sign recognition as an example, one possible application scope could be the set of all valid traffic signs erected in Germany as perceived by passenger cars in 2016.

Given a dataset that is representative for the intended application scope, the uncertainty in the model outcomes can be determined with a certain level of statistical confidence for this application scope. In real applications, however, test datasets are not always representative; for example, due to sample selection bias (which is a subcategory of dataset shift (Sugiyama, Lawrence, & Schwaighofer, 2009)), which falsifies the statistical results (Zadrozny, 2004).

Representative test datasets can be obtained by probabilistic sampling methods such as simple random sampling, sampling with unequal probability, systematic sampling, stratified random sampling, or cluster sampling (Nassiuma, 2000).

Despite the representativeness of the test dataset, statistically justified guarantees on previously determined uncertainties can only be provided if the model is applied within its intended application scope. Therefore, scope compliance is defined as the likelihood that a given data-driven model is currently applied to observations within its intended application scope (Kläs & Vollmer, 2018). For example, if a traffic sign recognition model tested with German traffic sign images is used on roads in the U.S., the model is applied outside its intended application scope.

Scope incompliance can be detected by monitoring known properties of the intended application scope and respective thresholds (i.e., known boundaries of generalizability). In addition, the crossing of boundaries that were unknown or not considered at design time may be detected at runtime using novelty detection techniques (Pimentel, Clifton, Clifton, & Tarassenko, 2014).

A specific type of scope incompliance that needs special consideration is concept drift. Concept drift describes the fact that the actual relationship between input and corresponding outcome, which the data-driven model is trying to capture, will change over time for any sufficiently complex setting that does not solely rely on stable, physical laws (Webb, Hyde, Cao, Nguyen, & Petitjean, 2016). Since any test dataset can only comprise data collected in the past, concept drift is unavoidable in the long term and must be considered for most data-driven models if applied in practice. For example, referring to traffic sign recognition, more than 40 traffic signs were newly introduced, changed, or deprecated in Germany in the year 2017. Methods addressing concept drift can be passive (i.e., they try to detect its occurrence (Harel, Mannor, El-Yaniv, & Crammer, 2014)) or active. Active methods adapt the model outcome, e.g., by retraining the model (Klinkenberg, 2004) or by using an ensemble of several models based on data from different time frames (Masud, Gao, Khan, Han, & Thuraisingham, 2009).

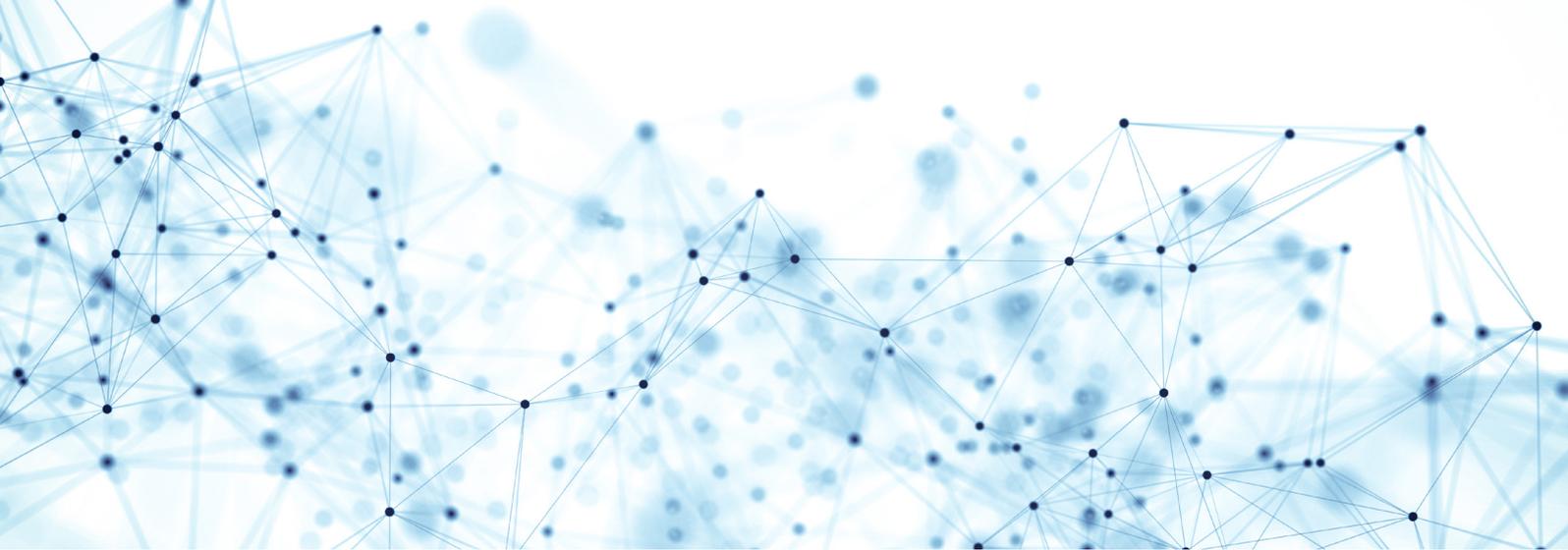

## Acknowledgments

Parts of this work are funded by the German Ministry of Education and Research (BMBF) under grant number 01IS16043E (CrESt).

# UNCERTAINTY IN AI AND MACHINE LEARNING – AN OVERVIEW

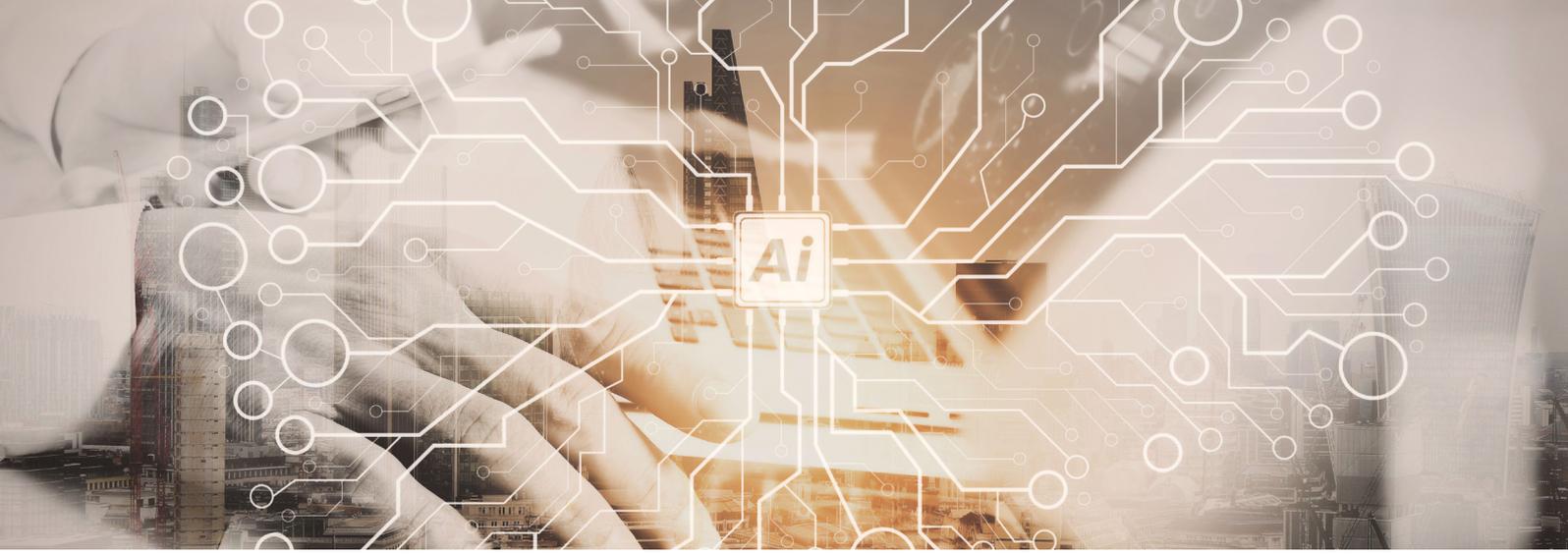